\documentclass[10pt,twocolumn,letterpaper]{article}

\usepackage[pagenumbers]{iccv} 

\definecolor{iccvblue}{rgb}{0.21,0.49,0.74}
\usepackage[pagebackref,breaklinks,colorlinks,allcolors=iccvblue]{hyperref}
\usepackage{bm}
\usepackage{url}
\usepackage{amsmath}
\usepackage{amssymb}
\usepackage{enumitem,kantlipsum}
\usepackage{graphicx}
\usepackage{subcaption}
\usepackage{bbm}
\usepackage{booktabs}
\usepackage{dsfont}
\usepackage{wrapfig}
\usepackage{makecell}
\usepackage[export]{adjustbox}

\def\paperID{3329} 
\def\confName{ICCV}
\def\confYear{2025}
\def\method{$\mathcal{D}$-Attn}

\title{\method{}: Decomposed Attention for Large Vision-and-Language Models}

\author{Chia-Wen Kuo \quad Sijie Zhu$^{*}$ \quad Fan Chen \quad Xiaohui Shen \quad Longyin Wen \\
ByteDance Intelligent Creation\\
}

\begin{document}
\maketitle
\let\thefootnote\relax\footnotetext{$*$ Corresponding author, sijiezhu@bytedance.com}
\begin{abstract}
Large vision-and-language models (LVLMs) have traditionally integrated visual and textual tokens by concatenating them into a single homogeneous input for large language models (LLMs), thereby maximally preserving the pre-trained language capabilities.
However, this constrained architecture for visual and textual tokens restricts the design space for processing visual tokens, potentially leading to suboptimal performance and efficiency.
In this paper, we propose Decomposed Attention (\method{}), a more flexible attention architecture for LVLMs, which enables modification of visual token operations without affecting textual-to-textual attention.
\method{} decomposes the 1-D causal self-attention of LVLMs into visual-to-visual, textual-to-visual, and textual-to-textual attentions, and the visual and textual output tokens from the decomposed attentions are merged with a carefully derived weighting strategy, namely $\alpha$-weighting. 
Taking advantage of the flexibility, we are able to introduce two critical improvements in visual token processing while maintaining the capacity of pre-trained LLMs: 1) We rectify the biased positional encoding in textual-to-visual attention to boost visual understanding performance. 2) We diagonalize visual-to-visual attention to reduce computation complexity from $\mathcal{O}(|V|^2)$ to $\mathcal{O}(|V|)$ for $|V|$ visual tokens without compromising performance. Extensive experiments and analysis validate the effectiveness of \method{}, demonstrating significant improvements on multiple image benchmarks while significantly reducing computational costs (\eg, $5\times$ faster). Code will be available at \url{https://github.com/bytedance/DecomposedAttention}.
\end{abstract}
\section{Introduction}\label{sec:intro}
\begin{figure}[tbp]
    \centering
    \includegraphics[width=.75\linewidth]{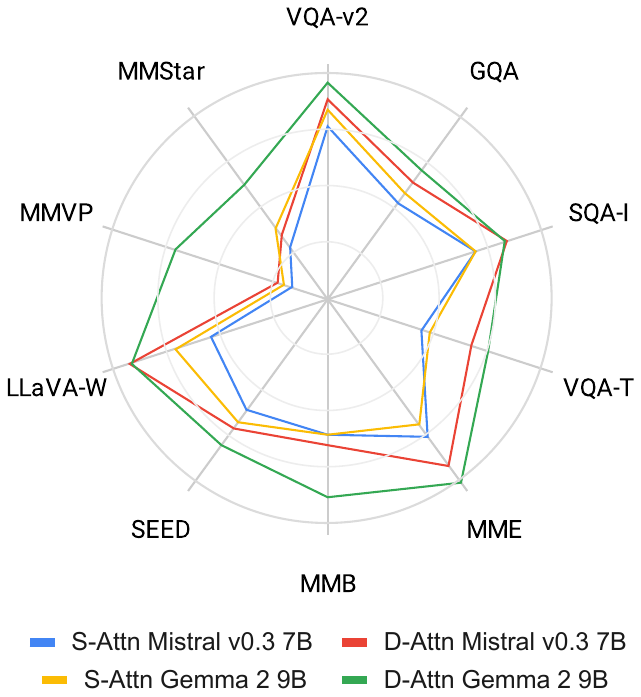}
    \caption{Performance comparison between proposed \method{} models and their self-attention (S-Attn) counterparts on popular image benchmarks. Detailed results are available in Table~\ref{tab:main}.}
    \label{fig:radar}
\end{figure}

Large Vision-and-Language Models (LVLMs) \citep{llava} have become pivotal in advancing artificial intelligence, and achieved significant advancement in various applications, \eg, image captioning, visual question answering, and multi-modal assistant. The success of LVLMs has largely benefited from the strong background knowledge and reasoning ability of pre-trained large language models (LLMs) like LLaMA~\citep{touvron2023llama,zheng2023judging} and Mistral~\citep{jiang2023mistral}. To leverage the capability of pre-trained LLMs, recent LVLMs~\citep{llava} usually concatenate textual and visual tokens equally as input of LLMs, so that the textual token operation within LLM is not affected. 
Given that visual tokens generated from vision encoders (\eg, CLIP~\citep{clip}) are inherently different from the textual tokens obtained from discrete tokenizers, this predominant way of handling them with one \textit{S-Attn} (Self-Attention) could be suboptimal in terms of performance and efficiency. 1) With the original positional encoding of LLMs, textual tokens intrinsically attend more to the visual tokens with closer position in the concatenated token sequence, which is biased and undesired, because different visual tokens are equal for textual tokens. 2) The computation complexity of the self-attention between visual tokens is $O(|V|^{2})$ ($|V|$ denotes the number of visual tokens), which is highly redundant, as the visual tokens are generated from bi-directional transformers and their correlation information has been encoded. \textit{However, the current S-Attn architecture lacks flexibility to improve on these aspects.} 

In contrast, another line of LVLMs~\citep{awadalla2023openflamingo,dubey2024llama} equip \textit{X-Attn} (Cross-Attention) which offers more flexibility for specific design of visual tokens operation, \eg, incorporating visual tokens into an LLM via tailored modules or operations such as a cross-attention block or a \texttt{Tanh} gate. However, these architectural or operational changes could break the integrity of pre-trained LLMs  and lead to inferior VL performance~\citep{flamingo}. \textit{It remains unknown on how to maintain capability of LLMs to match the performance of S-Attn.}

\begin{figure}[tbp]
\centering
\includegraphics[width=0.65\linewidth]{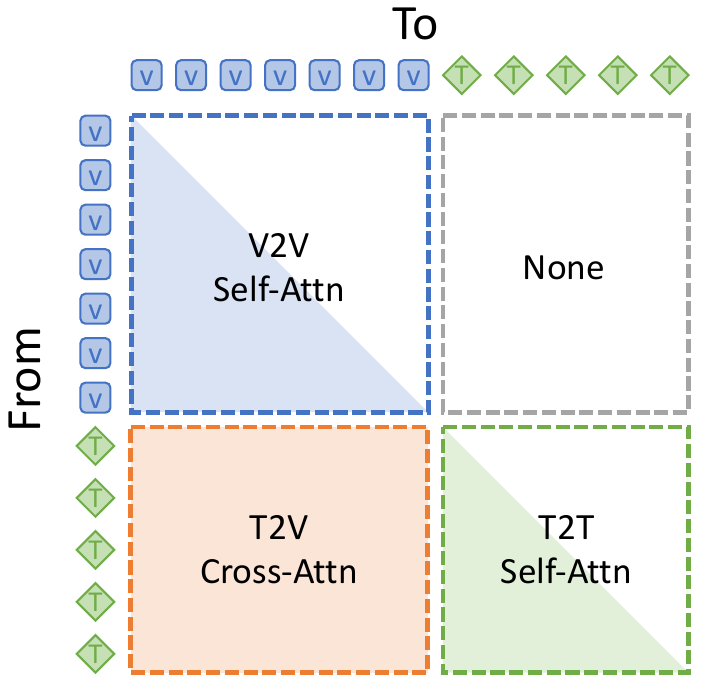}
\caption{Decomposition of causal self-attention within an LVLM into visual-to-visual self-attention (\textcolor{blue}{V2V Self-Attn}), textual-to-visual cross-attention (\textcolor{orange}{T2V Cross-Attn}), and textual-to-textual self-attention (\textcolor{green}{T2T Self-Attn}).}\label{fig:attn}
\end{figure}

In this paper, we propose \textbf{Decomposed Attention (\method{})}, a novel attention architecture to bridge the gap between S-Attn and X-Attn, \ie, preserving the performance advantage of S-Attn while enjoying the design flexibility of X-Attn.
By decomposing the causal self-attention into V2V, T2V, and T2T attentions (Figure~\ref{fig:attn}), \method{} gains the flexibility to tailor visual-related operations for better visual information incorporation into LLMs, while maximally preserving the LLM's pre-trained capability with a carefully derived merging strategy (\ie, alpha weighting strategy) between visual and textual tokens. In Sec.~\ref{sec:debiased-pos-enc}, we propose \textbf{debiased positional encodings} to eliminate the undesirable positional bias between visual and textual tokens such that the LLM can better leverage visual information, bringing substantial improvements on visual understanding capability. In Sec.~\ref{sec:v2v-diag}, we propose \textbf{V2V diagonalization} to substantially reduce the computational complexity from $\mathcal{O}(|V|^2)$ to $\mathcal{O}(|V|)$ for $|V|$ visual tokens without sacrificing performance. Extensive experiments on 10 VQA benchmarks~\citep{llava} and multiple LLM backbones~\citep{jiang2023mistral,team2024gemma} demonstrate the superior performance (Figure~\ref{fig:radar}) and remarkable efficiency (Table \ref{tab:ablation-arch}) of the proposed \method{}. Contributions are summarized as follows:
\begin{itemize}
    \item We propose a novel Decomposed Attention, which maintains the language capability of pre-trained LLMs and offers flexibility for processing of visual tokens.
    \item We introduce debiased positional encoding so that the textual tokens are not biased toward certain visual tokens.
    \item We introduce diagonized visual-to-visual attention to reduce the computation from $O(|V|^2)$ to $O(|V|)$, where $|V|$ denotes the number of visual tokens.
    \item Extensive experiments demonstrating siginicant performance boost and much less (\eg, $5\times$ faster) computational cost.
\end{itemize}



\section{Related Work}\label{sec:related}

Emerging large vision-language models (LVLMs) have made significant progress in visual understanding, particularly in Visual Question Answering (VQA).
The predominant architectures can be summarized as different combinations of a vision encoder, an adapter, and a large language model (LLM).
To name a few, LLaVA \citep{llava}, LLaVA NeXT \citep{llava_next}, LLaVA-OneVision \citep{llava_onevision}, Instruct BLIP \citep{dai2023instructblip}, BLIP3 \citep{blip3}, VILA \citep{lin2024vila}, Qwen2-VL \citep{wang2024qwen2}, CuMo \citep{li2024cumo}, Intern-XC-2.5 \citep{zhang2024internlm}, miniGemini \citep{li2024mini}, Cambrian-1 \citep{tong2024cambrian}, Phi-3 VL \citep{abdin2024phi}, Chameleon~\citep{team2024chameleon}, Molmo \citep{deitke2024molmo}, Phi-3.5-Vision~\citep{abdin2024phi}.
Despite differences in data, vision encoders, or adapter, all these works adhere to a decoder-only LLM architecture that process visual and textual tokens equally within an LLM using the causal self-attention mechanism \citep{vaswani2017attention}.
This line of works introduces minimal changes to the LLM, thus maximally preserving the pre-trained capability of an LLM and resulting in superior performance.
\textit{Nevertheless, S-Attn has limited flexibility to process visual tokens with specifically tailored designs within an LLM, as modifications could degrade the language capability of the pre-trained LLM.}

In contrast to predominant LVLM architectures, models like Flamingo~\citep{flamingo,awadalla2023openflamingo}, IDEFICS~\citep{laurenccon2024obelics}, and LLaMA 3~\citep{dubey2024llama} integrate visual information into LLMs via cross-attention mechanisms between textual and visual tokens.
These architectures share similarities with our proposed \method{}, such as employing text-to-vision cross-attention to incorporate visual information and achieving a computational complexity of $\mathcal{O}(|V|)$ for $|V|$ visual tokens.
However, this line of works introduce significant architectural and operational changes within the pre-trained LLM, such as appending additional cross-attention modules, or introducing \texttt{Tanh} or \texttt{Sigmoid} gating.
These substantial changes can compromise the integrity of the pre-trained LLM, potentially degrading its inherent capabilities.
Indeed, \cite{laurenccon2024matters} show in IDEFICS-2 that X-Attn architectures underperform decoder-only S-Attn architectures, leading them to discard the X-Attn design.
Despite its inferior performance, X-Attn offers more flexibility to process visual tokens differently within an LLM.
\textit{Therefore, we could potentially enhance an LVLM's visual understanding capability by properly designing the operations for incorporating visual information.}


\section{Method}\label{sec:method}



In modern LVLMs, visual inputs are encoded by a visual encoder (\textit{e.g.} CLIP or SigLip) into a sequence of tokens.
These visual tokens are then concatenated with a sequence of text tokens (\textit{e.g.} questions or prompts) into a long unified sequence, and then fed into a pre-trained LLM for joint processing.
To better understand how visual tokens are processed within an LLM, we begin by decomposing the causal self-attention mechanism in an LLM when both visual and textual tokens are present.
Assuming visual tokens are placed in front of textual tokens, as illustrated in Figure~\ref{fig:attn}, causal self-attention can be split into three distinct components: (1) visual-to-visual self-attention (V2V SA), (2) textual-to-visual cross-attention (T2V XA), and (3) textual-to-textual self-attention (T2T SA).

We can see that visual tokens interacts with textual tokens in T2V XA.
Visual tokens are also processed by the LLM via V2V SA.
In Section~\ref{sec:debiased-pos-enc}, we identify an undesirable positional bias between visual and textual tokens in T2V XA and propose a debiased postional encodings to improve the model's visual understanding capability.
In Section~\ref{sec:v2v-diag}, we propose V2V Diagonalization to reduce the computational complexity from $\mathcal{O}(|V|^2)$ to $\mathcal{O}(|V|)$ without sacrificing model performance.
Finally, in Section~\ref{sec:decomp-attn}, we derive how a causal self-attention can be equivalently decomposed in the way shown in Figure~\ref{fig:attn}.
This decomposition offers the flexibility that enable us to introduce our proposed processing of visual tokens within the LLM.
Also crucially, this decomposition maintains equivalence to LLM's innate causal self-attention mechanism, thus maximally preserving its pre-trained capability.

\subsection{Debiased Positional Encodings}\label{sec:debiased-pos-enc}

\begin{figure}[!htbp]
\vspace{-0.4cm}
\centering
\begin{subfigure}{.48\linewidth}
  \includegraphics[width=1.05\linewidth,right]{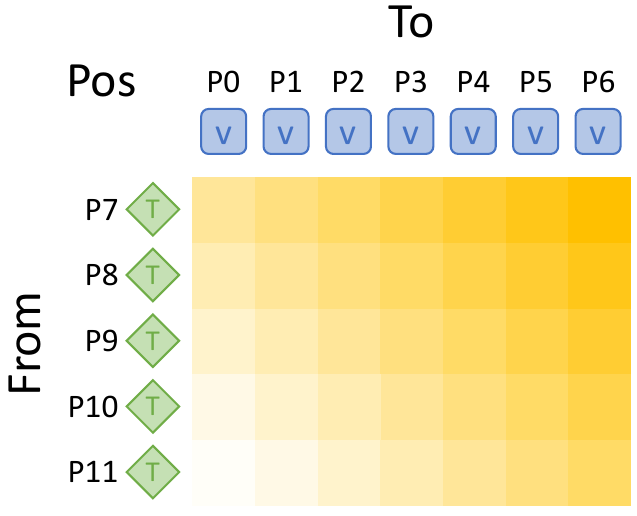}
  \caption{The biased positional encodings between visual and textual tokens in T2V XA, where each textual token has lower attention weight bias (lighter color) toward further visual tokens.}\label{fig:pos-bias}
\end{subfigure}
\hfill
\begin{subfigure}{.48\linewidth}
  \includegraphics[width=1.05\linewidth,right]{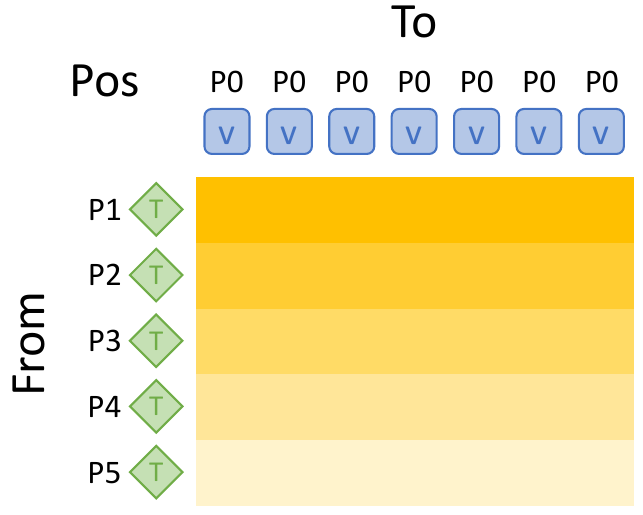}
  \caption{The proposed debiased positional encodings between visual and textual tokens in T2V XA, where each textual tokens has uniform attention weight bias toward all visual tokens.}\label{fig:pos-debias}
\end{subfigure}
\vspace{-0.3cm}
\caption{An illustration of positional encoding.}
\end{figure}

In T2V XA, textual tokens interact with visual tokens to incorporate visual information, where textual tokens (queries) attend to visual tokens (keys and values) as shown in Figure~\ref{fig:pos-bias}.
In the figure, we can see that the rotary/relative positional encodings skew attention weights based on the positional distance between visual and textual tokens.
For example, distant pairs such as the textual token at P9 and visual token at P0 receive lower attention weight than pairs closer together, like the textual token at P9 and visual token at P6.
Since the 2D visual tokens are typically flattened in a row major way, this attention pattern equivalently means that a textual token will pay more attention to the bottom of an image than the top of an image.
This bias can hinder effective vision-language interaction for tasks requiring a comprehensive understanding of visual contents.

To address this issue, we propose to debias T2V XA by setting the position of visual tokens to the same value.
For example, as shown in Figure~\ref{fig:pos-debias}, we set the position of all visual tokens to P0.
In this way, each textual token now has uniform attention weight biases toward all visual tokens, enhancing the comprehensive understanding of visual contents and hence resulting in stronger performance.

\begin{figure*}[!htbp]
\centering
\begin{subfigure}{.24\linewidth}
  \centering
  \includegraphics[width=1\linewidth]{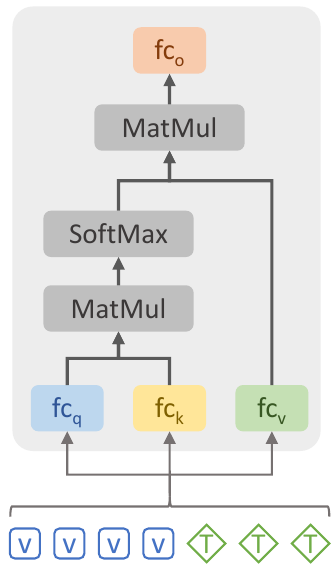}
  \caption{Conventional S-Attn.}\label{fig:arch-attn}
\end{subfigure}
\hfill
\begin{subfigure}{.24\linewidth}
  \centering
  \includegraphics[width=1\linewidth]{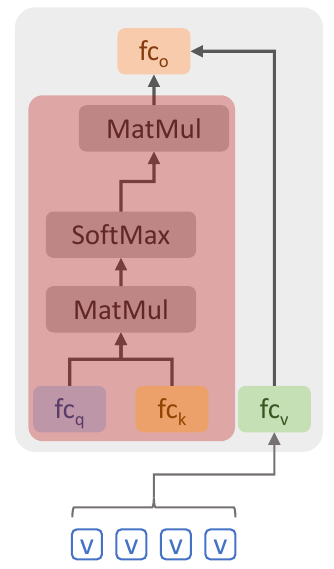}
  \caption{Diagonalized V2V attention.}\label{fig:arch-diag}
\end{subfigure}
\hfill
\begin{subfigure}{.45\linewidth}
  \centering
  \includegraphics[width=1\linewidth]{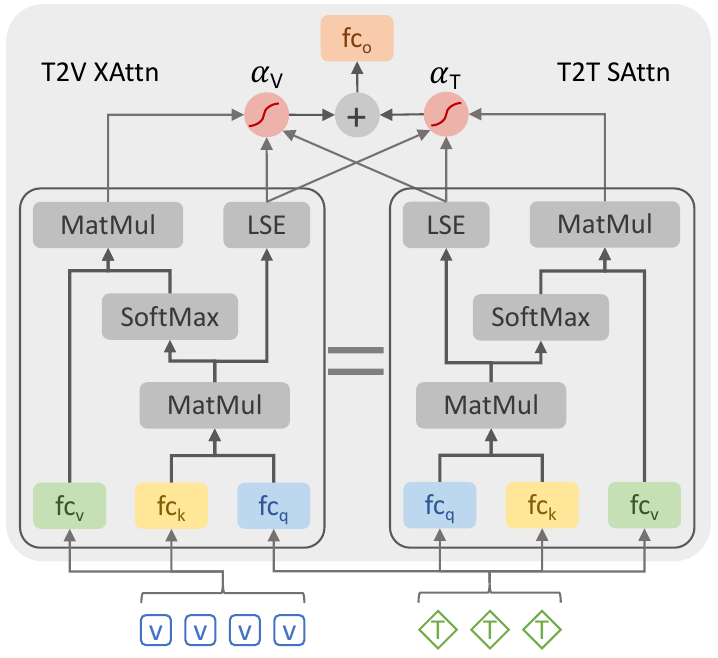}
  \caption{The $\alpha$-weighting strategy to merge T2V X-Attn and T2T S-Attn.}\label{fig:arch-dcmp}
\end{subfigure}
\vspace{-0.3cm}
\caption{
Module architecture and operations of
(a) conventional attention in LVLM with visual and texual tokens concatenated as a homogeneous input sequence,
(b) V2V Diagonal-Attn, where the expensive computation of softmax attention weight is skipped, and
(c) $\alpha$-weighting strategy to merge T2V Cross-Attn and T2T Self-Attn equivalent to LVLM's inherent attention operations for retaining a pre-trained LLM's full capability.
}\label{fig:arch}
\end{figure*}

\subsection{V2V Diagonalization }\label{sec:v2v-diag}

In V2V SA, each visual tokens attend to all other visual tokens in a causal way to model the pairwise interaction between visual tokens.
Given a sequence of $|V|$ visual tokens, the computational complexity of V2V SA is $\mathcal{O}(|V|^2)$, which could be a problem for processing high-resolution images or even videos with hundreds of frames.

Motivated by our observation that the attention weights matrix of V2V SA are concentrated at the diagonal, indicating that each visual token primarily looks at itself, we propose to diagonalize V2V SA to significantly reduce computational complexity.
In V2V diagonalization, each visual token attends only to itself, rather than to all other visual tokens, such that the computational complexity could be reduced from $\mathcal{O}(|V|^2)$ to $\mathcal{O}(|V|)$.
Mathematically, for visual tokens $V \in |V|\times d$:
\begin{align}
\bar{V} &= \texttt{SA}(V, V) \nonumber\\ &= \texttt{fc}_o\left( \underbrace{\texttt{softmax}\left(\frac{\texttt{fc}_q(V)\ \texttt{fc}_k(V)^T}{\sqrt{d}}\right)}_{\text{diagonalize} }\texttt{fc}_v(V) \right) \nonumber\\
&\Rightarrow \texttt{fc}_o\left(\mathds{1}\ \texttt{fc}_v(V) \right) =  \texttt{fc}_o\left(\texttt{fc}_v(V) \right)\label{eq:diag-attn}
\end{align}
, where $\mathds{1}$ is an identity matrix of size $|V|\times |V|$, and $\texttt{fc}_q$, $\texttt{fc}_k$, $\texttt{fc}_v$, and $\texttt{fc}_o$ are fully connected layers in an attention module for query, key, value, and output, respectively.

By turning the softmax attention matrix into an identity matrix, we essentially force each visual token to only attend to itself, bypassing the need for pairwise interactions between visual tokens.
As shown in Equation~\ref{eq:diag-attn} and Figure~\ref{fig:arch-diag}, this diagonalization simplifies the self-attention operation to only two fully connected layers, thus significantly reducing the computational complexity from $\mathcal{O}(|V|^2)$ to $\mathcal{O}(|V|)$ for $|V|$ visual tokens.
V2V Diagonal-Attn is particularly valuable when dealing with high-resolution images or long video inputs, where the number of visual tokens $|V|$ becomes large. 
Notably, in our experiments, we demonstrate that this method achieves similar performance to full attention while offering significant computational savings.

\subsection{Decomposed Attention}\label{sec:decomp-attn}

The debiased positional encodings and V2V diagonalization proposed above for enhancing performance and efficiency are based on the assumption that causal self-attention mechanism can be decomposed as shown in Figure~\ref{fig:attn}.
In this section, we mathematically derive how we can decompose into three blocks: V2V SA, T2V XA, and T2T SA; and how these blocks interact.

As illustrated in Fig.~\ref{fig:attn}, a visual token $v$ only attends to other visual tokens $V$.
Therefore, the attention output $\bar{v}$ for a visual token is simply:
\begin{equation}
\bar{v} = \texttt{Attn}(v, V) = \texttt{SA}(v, V) \label{eq:v2v-sa}
\end{equation}

On the other hand, as we can see in Figure~\ref{fig:attn}, a textual token $t$ attends to \textbf{both} visual tokens $V$ and textual tokens $T$, indicating the interplay between T2V XA and T2T SA.
Mathematically, the attention output $\bar{t}$ for a textual token can be expressed as:
\begin{equation}
\bar{t} = \texttt{Attn}(t, [V, T]) = \sum_{i}^{L} \frac{e^{\bm{q}_t\cdot \bm{k}_{i}}}{\sum_{l}^{L} e^{\bm{q}_{t}\cdot \bm{k}_{l}}}\bm{v}_{i} \label{eq:xattn}
\end{equation}
, where $\bm{q}, \bm{k}, \bm{v}$ are projected query, key, value within an attention module, respectively.
$\bm{k}$ and $\bm{v}$ are projected from the concatenated visual and textual tokens $[V, T]$.
For $N$ visual tokens and $M$ textual tokens, $\bm{k}_i \in \{\bm{k}_{v_1},...,\bm{k}_{v_N},\bm{k}_{t_1},...,\bm{k}_{t_M},\}$, where $\bm{k}_{v_j}$ and $\bm{k}_{t_l}$ represent the key corresponding to the $j$-th visual token and $l$-th textual tokens, respectively. 
Similarly $\bm{v}_i \in \{\bm{v}_{v_1},...,\bm{v}_{v_N},\bm{v}_{t_1},...,\bm{v}_{t_M}\}$.
We then rewrite Equation~\ref{eq:xattn} by splitting key value from $V$ and from $T$:
\begin{equation*}
\sum_{i}^{L} \frac{e^{\bm{q}_t\cdot \bm{k}_{i}}}{\sum_{l}^{L} e^{\bm{q}_{t}\cdot \bm{k}_{l}}}\bm{v}_{i}
= \underbrace{\sum_{i}^{N} \frac{e^{\bm{q}_t\cdot \bm{k}_{v_i}}}{\sum_{l}^{L} e^{\bm{q}_{t}\cdot \bm{k}_{l}}}\bm{v}_{v_i}}_{(a)} +
\underbrace{\sum_{i}^{M} \frac{e^{\bm{q}_t\cdot \bm{k}_{t_i}}}{\sum_{l}^{L} e^{\bm{q}_{t}\cdot \bm{k}_{l}}}\bm{v}_{t_i}}_{(b)}
\end{equation*}
The term $(a)$ can be further derived as:
\begin{align}
&\underbrace{\sum_{i}^{N} \frac{e^{\bm{q}_t\cdot \bm{k}_{v_i}}}{\sum_{l}^{L} e^{\bm{q}_{t}\cdot \bm{k}_{l}}}\bm{v}_{v_i}}_\text{(a)} =\frac{\sum_{n}^{N} e^{\bm{q}_{t}\cdot \bm{k}_{v_n}}}{\sum_{l}^{L} e^{\bm{q}_{t}\cdot \bm{k}_{l}}} \sum_{i}^{N} \frac{e^{\bm{q}_t\cdot \bm{k}_{v_i}}}{\sum_{n}^{N} e^{\bm{q}_{t}\cdot \bm{k}_{v_n}}}\bm{v}_{v_i}\nonumber\\
&=\frac{\sum_{n}^{N} e^{\bm{q}_{t}\cdot \bm{k}_{v_n}}}{\sum_{l}^{L} e^{\bm{q}_{t}\cdot \bm{k}_{l}}} \texttt{XA}(t,V)
\equiv \alpha_V\ \texttt{XA}(t, V)
\end{align}
Similarly, for the term $(b)$:
\begin{align}
&\underbrace{\sum_{i}^{M} \frac{e^{\bm{q}_t\cdot \bm{k}_{t_i}}}{\sum_{l}^{L} e^{\bm{q}_{t}\cdot \bm{k}_{l}}}\bm{v}_{t_i}}_\text{(b)}=\frac{\sum_{m}^{M} e^{\bm{q}_{t}\cdot \bm{k}_{t_m}}}{\sum_{l}^{L} e^{\bm{q}_{t}\cdot \bm{k}_{l}}} \sum_{i}^{M} \frac{e^{\bm{q}_t\cdot \bm{k}_{t_i}}}{\sum_{m}^{M} e^{\bm{q}_{t}\cdot \bm{k}_{t_m}}}\bm{v}_{t_i}\nonumber\\
&=\frac{\sum_{m}^{M} e^{\bm{q}_{t}\cdot \bm{k}_{t_m}}}{\sum_{l}^{L} e^{\bm{q}_{t}\cdot \bm{k}_{l}}} \texttt{SA}(t,T)
\equiv \alpha_T\ \texttt{SA}(t,T)
\end{align}

For numerical stability in modern deep learning frameworks, we take log of the summed exponentials as:
\begin{equation*}
\text{Let}\ S_V = \log\left(\sum_{n}^{N} e^{\bm{q}_{t}\cdot \bm{k}_{v_n}}\right) \text{, }\ S_T = \log\left(\sum_{m}^{M} e^{\bm{q}_{t}\cdot \bm{k}_{t_m}}\right)
\end{equation*}
Then the weights $\alpha_V$ can be expressed as:
\begin{align}
\alpha_V
&= \frac{\sum_{n}^{N} e^{\bm{q}_{t}\cdot \bm{k}_{v_n}}}{\sum_{l}^{L} e^{\bm{q}_{t}\cdot \bm{k}_{l}}}
= \frac{e^{S_V}}{e^{S_V}+e^{S_T}}
= \frac{1}{1+e^{-(S_{V}-S_{T})}}\nonumber\\
&= \texttt{Sigmoid}(S_V - S_T) \label{eq:alpha-v}
\end{align}
We can similarly derive the weights $\alpha_T$ as:
\begin{equation}
\alpha_T = \texttt{Sigmoid}(S_T - S_V) = 1-\alpha_V \label{eq:alpha-t}
\end{equation}
In summary, the attention output $\bar{t}$ for an input textual token can be expressed as:
\begin{equation}
\bar{t} = \texttt{Attn}(t, [V, T]) = \alpha_V\ \texttt{XA}(t, V) + \alpha_T\ \texttt{SA}(t,T) \label{eq:decomp-attn}
\end{equation}
, a weighted sum of T2V XA and T2T SA, where weights $\alpha_V$ and $\alpha_T$ can be analytically computed from Equation \ref{eq:alpha-v} and \ref{eq:alpha-t}, respectively. This merging strategy is abbreviated as \textbf{$\alpha$-weighting strategy}.\\

With the decomposition of causal self-attention derived above, we can seamlessly incorporate debiased positional encodings and V2V diagonalization proposed in Section~\ref{sec:debiased-pos-enc} and \ref{sec:v2v-diag}, respectively.
Specifically, we can apply debiased positional encodings in T2V XA and then combine it with T2T SA following Equation~\ref{eq:decomp-attn}.
We can also apply V2V diagonalization in V2V SA following Equation~\ref{eq:v2v-sa}.
The decompostion offers us the flexibility to process visual and textual tokens differently within an LLM.
In addition to the flexibility offered by the decomposition derived above, \method{} does not introduce architectural and operational changes to the causal self-attention mechanism.
\textit{This is crucial for preserving the pre-trained LLM's capability, leading to superior visual understanding performance.}

\section{Experiment}
\label{sec:experiment}

\subsection{Implementation Details}


\noindent\textbf{Model:} Our proposed \method{} model is built based on the architecture of LLaVA~\citep{llava}.
It is constructed using three primary components: a pre-trained SigLip~\citep{zhai2023sigmoid} visual encoder, a randomly initialized two-layer MLP adapter with RMSNorm~\citep{zhang2019root}, and a pre-trained LLM.
We modify only the decoder layer and self-attention mechanisms within the LLM to implement our \method{}.
In this paper, we experiment with two different LLM families: Mistral v0.3 7B~\citep{jiang2023mistral}, and Gemma 2 9B~\citep{team2024gemma}.



\noindent\textbf{Training:} The training of \method{} follows a three-stage strategy outlined in ShareGPT4V~\citep{chen2023sharegpt4v}.
In the first stage, the MLP adapter is pre-trained on LLaVA's LAION/CC/SBU\citep{llava,schuhmann2022laion,sharma2018conceptual,ordonez2011im2text} 58k for modality alignment.
In the second stage, the entire model is fine-tuned using 1.25M dense captions from the ShareGPT4V-PT dataset~\citep{chen2023sharegpt4v}.
In the third and final stage, we perform instruction tuning using a combined dataset of 665k examples from LLaVA-1.5~\citep{llava} and 102k dense captions from ShareGPT4V~\citep{chen2023sharegpt4v}.
The entire training procedure completes in 24 hours on 32 H100 GPUs with DeepSpeed \cite{rasley2020deepspeed}.
Detailed hyperparameters are provided in the Appendix.


\noindent\textbf{Evaluation:} Following LLaVA's evaluation protocol, we evaluate \method{} on ten image benchmarks, including VQA-v2~\citep{goyal2017making}, GQA~\citep{hudson2019gqa}, SQA-I~\citep{lu2022learn}, VQA-T~\citep{mao2016generation}, MME~\citep{fu2024mmecomprehensiveevaluationbenchmark}, MMB~\citep{liu2023mmbench}, SEED-I~\citep{li2023seed}, LLaVA-W~\citep{llava}, MMVP~\citep{tong2024eyes}, and MMStar~\citep{chen2024we}.

\begin{table*}[!htbp]
\centering
\renewcommand{\arraystretch}{1.2}
\resizebox{1.0\linewidth}{!}{
\begin{tabular}{@{\extracolsep{4pt}}lcc|cccccccccccc@{}}
\toprule
Method & LLM & Data & VQA-v2 & GQA & SQA-I & VQA-T & MME & MMB & SEED-I & LLaVA-W & MMVP & MMStar \\
\midrule
InstructBLIP & Vicuna 7B & 130.2M & - & 49.2 & 60.5 & 50.1 & - & 36.0 &  60.5 &  60.9 & - & - \\
BLIP 3 & Phi 3 3.8B & 3T & - & - & 88.3 & 71.0 & 1288.0 & 76.8 & 72.2 & - & - & 48.1 \\
VILA & LLaMA 2 7B & 51M & 79.9 & 62.3 & 68.2 & 64.4 & 1533.0 & 68.9 & 61.1 & 69.7 & - & - \\
IDEFICS & LLaMA 7B & 354M & 50.9 & 38.4 & - & 25.9 & - &  48.2 & - & - & - & - \\
Mini-Gemini & LLaMA 3 8B & 9.5M & - & 64.5 & 75.1 & 70.2 & 1606.0 & 65.8 & 73.7 & - & 18.7 & -  \\
Cambrian & LLaMA 3 8B & 9.5M & - & 64.6 & 80.4 & 71.7 & 1547.1 & 75.9 & 74.7 & - & 51.5 & - \\
Qwen-VL & Qwen 7B & 1.4B & 78.8 & 59.3 & 67.1 &  63.8 & - &  38.2 & 56.3 & - & - & - \\
Qwen2-VL & Qwen 2 7B & UNK & - & - & - & 84.3 & - & 83.0 & - & - & - & 60.7 \\
Intern-XC & InternLM 7B & 1.1B & - & - & - & - & 1528.4 & 74.4 & 66.9 & - & - & -  \\
Intern-XC 2.5 & InternLM 2 7B & UNK & - & - & - & 78.2 & - & 82.2 & 75.4 & - & - & 59.9 \\
CuMo & Mistral 0.2 7B & 2.9M & 82.2 & 64.9 & 73.9 & 67.0 & 1548.6 & 73.0 & 72.1 & 85.7 & - & - \\
LLaVA-1.5 & Vicuna 1.5 7B & 1.2M & 78.5 & 62.0 & 66.8 & 58.2 & 1510.7 & 64.3 & 66.1 & 63.4 & 20.0 & 32.8 \\
LLaVA-1.6 & Mistral 0.2 7B & 1.4M & 82.2 & 64.8 & 72.8 & 65.7 & 1498.0 & 68.7 & 72.2 & 83.2 & 32.0 & 36.1 \\
\midrule
S-Attn & Mistral 0.3 7B & 2.5M & 80.3 & 61.8 & 72.7 & 62.2 & 1533.1 & 
70.3 & 70.5 & 70.7 & 28.0 & 36.8 \\
\method{} & Mistral 0.3 7B & 2.5M & 82.9 & 64.4 & 75.7 & 68.3 & 1598.6 & 71.3 & 72.6 & 79.8 & 30.0 & 38.3 \\
S-Attn & Gemma 2 9B & 2.5M & 81.8 & 63.0 & 72.8 & 63.2 & 1506.6 & 70.3 & 71.9 & 74.6 & 29.3 & 39.2 \\
\method{} & Gemma 2 9B & 2.5M & 84.3 & 65.9 & 75.5 & 70.7 & 1636.7 & 76.5 & 74.6 & 79.7 & 45.3 & 45.0 \\
\bottomrule
\end{tabular}
}
\caption{
Main results on a range of popular image benchmarks for our \method{} models, their S-Attn counterparts, and other SoTA models.
}\label{tab:main}
\vspace{-0.2cm}
\end{table*}
\begin{table*}[t]
\centering
\resizebox{1.\linewidth}{!}{
\begin{tabular}{@{\extracolsep{4pt}}cc|c|c|ccccccc@{}}
\toprule
Diag. Attn & Debiased Pos. & max $|V|$ $\uparrow$ & sec / it $\downarrow$ & GQA & VQA-T & MME & MMB & SEED-I & LLaVA-W & MMStar \\
\midrule
N & N & 9k & 11.25  & 61.8 & 62.2 & 1533.1 & 70.3 & 70.5 & 70.7 & 36.8 \\
Y & N & \textbf{74k} & \textbf{2.24} & 63.4 & 63.4 & 1507.6 & 68.8 & 70.7 & 71.2 & 32.6 \\
Y & Y & \textbf{74k} & \textbf{2.24} & \textbf{64.4} & \textbf{68.3} & \textbf{1598.6} & \textbf{71.3} & \textbf{72.6} & \textbf{79.8} & \textbf{38.3} \\
\bottomrule
\end{tabular}
}
\caption{
Ablations on V2V Diagonal-Attn and debiased positional encodings.
}\label{tab:ablation-arch}
\end{table*}

\begin{table*}[!htbp]
\centering
\begin{tabular}{@{\extracolsep{4pt}}c|c|cccccccc@{}}
\toprule
Merging Strategy & \#Params & GQA & SQA-I & VQA-T & MME & MMB & SEED-I & LLaVA-W & MMStar \\
\midrule
Cascade & 9.0 B & 64.1 & 72.9 & 67.0 & 1586.1 & 71.1 & 71.8 & 76.0 
& 36.4 \\
Tanh & \textbf{7.6 B} & 56.6 & 73.4 & 50.0 & 1337.6 & 62.4 & 59.4 & 59.3 & 33.3 \\
Sigmoid & \textbf{7.6 B} & \textbf{64.7} & 71.9 & 66.8 & 1548.1 & 69.2 & 72.4 & 73.2 & 35.8 \\
$\alpha$-weighting (ours) & \textbf{7.6 B} & 64.4 & \textbf{75.7} & \textbf{68.3} & \textbf{1598.6} & \textbf{71.3} & \textbf{72.6} & \textbf{79.8} & \textbf{38.3} \\
\bottomrule
\end{tabular}
\caption{
Ablations on various strategies for merging visual and textual tokens.
}\label{tab:ablation-merge}
\vspace{-0.2cm}
\end{table*}

\begin{figure*}[!htbp]
\centering
\includegraphics[width=1.\linewidth]{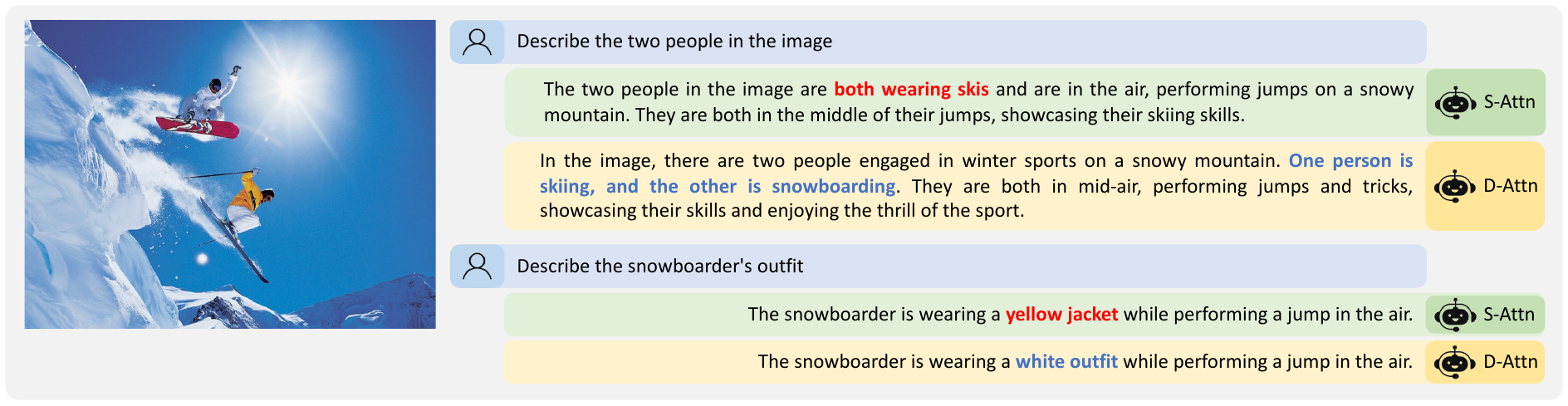}
\vspace{-0.2cm}
\includegraphics[width=1.\linewidth]{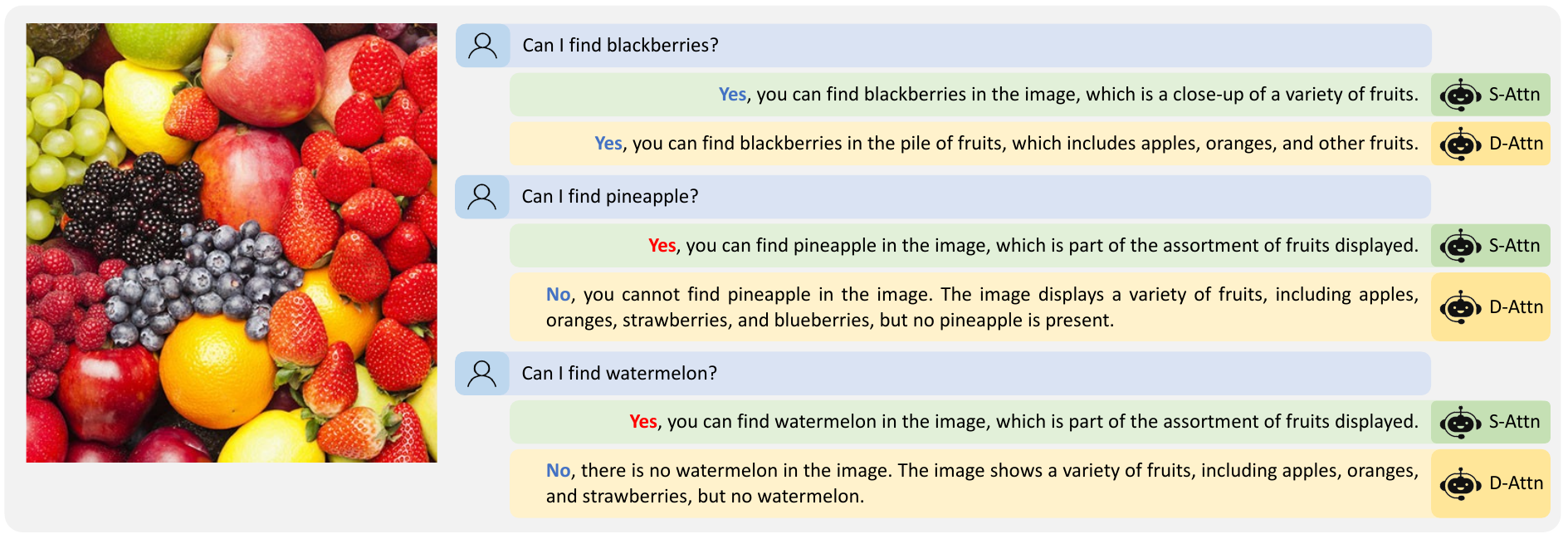}
\vspace{-0.2cm}
\includegraphics[width=1.\linewidth]{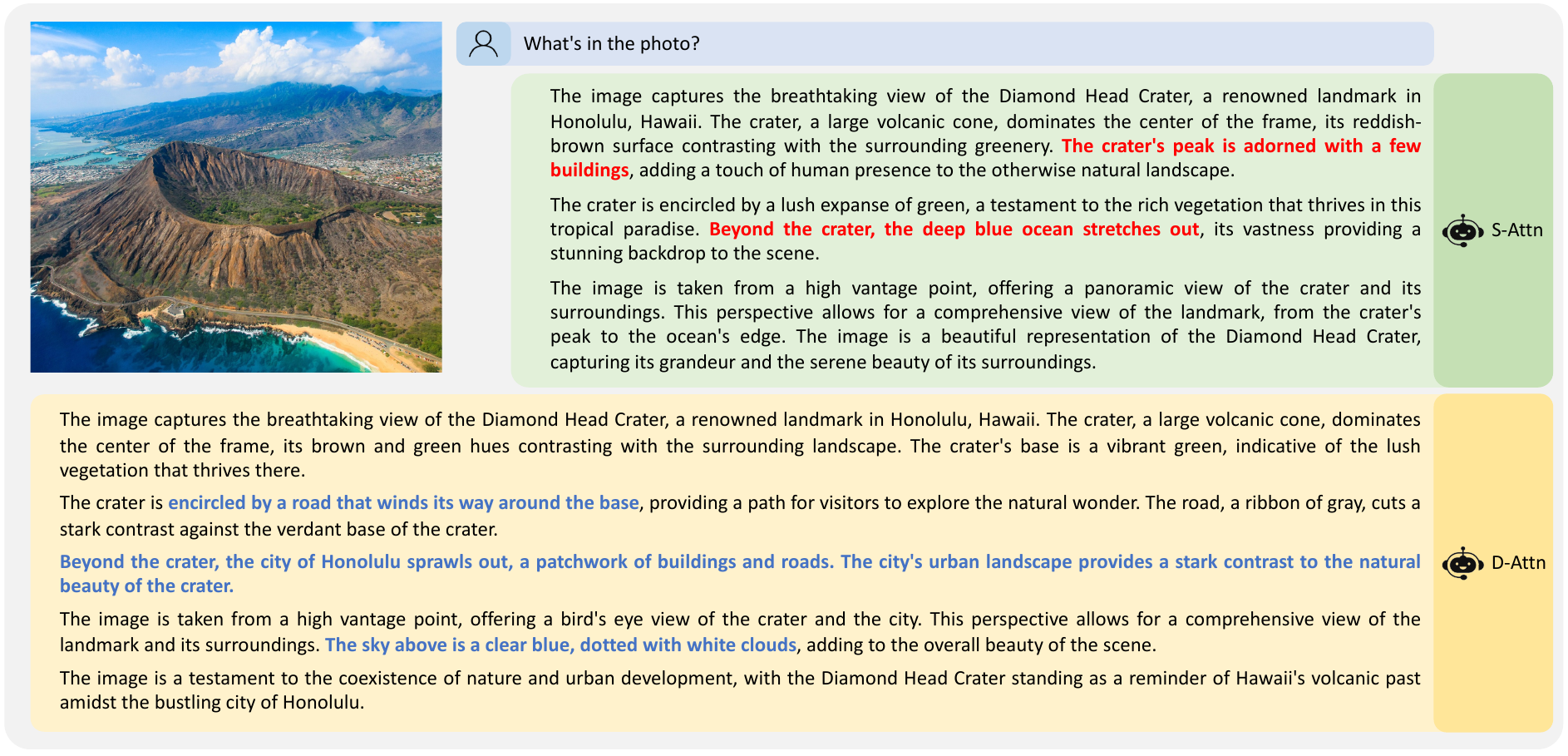}
\caption{
Qualitative comparisons between \method{} and its Self-Attn (S-Attn) counterpart. Erroneous outputs from the S-Attn model are highlighted in red, while the accurate and preferred responses from \method{} are highlighted in blue. 
}\label{fig:qualitative}
\end{figure*}

Our primary objective is not to achieve state-of-the-art performance but to rigorously validate the effectiveness of our proposed \method{} framework.
To ensure fair comparisons and facilitate reproducibility, we train \method{} using only publicly available datasets through supervised fine-tuning and construct the model with open-source pre-trained LLMs and visual encoders.
For stronger performance, researchers may scale up training data and models or apply more advanced training techniques such as Reinforcement Learning from Human Feedback (RLHF)\citep{bai2022training} or Direct Preference Optimization (DPO)\citep{rafailov2024direct}, which we leave as future work.


\subsection{Main Results}

We conduct experiments using Gemma 2 9B and Mistral v0.3 7B LLMs.
To ensure a fair comparison, both \method{} and S-Attn models are trained on the same datasets using identical training strategies and are constructed with the same pre-trained visual encoders and LLMs.


Table~\ref{tab:main} presents the results of our \method{} models and their S-Attn counterparts alongside other state-of-the-art LVLMs on ten popular image benchmarks.
For reference, we include models such as Instruction BLIP~\citep{dai2023instructblip}, BLIP3~\citep{blip3}, VILA~\citep{lin2024vila}, IDEFICS~\citep{laurenccon2024obelics},  Mini-Gemini~\citep{li2024mini}, Cambrian~\citep{tong2024cambrian}, Qwen-VL / Qwen2-VL~\citep{wang2024qwen2}, Intern-XC / Intern-XC 2.5~\citep{zhang2024internlm}, CuMo~\citep{li2024cumo}, and LLaVA-1.5 / LLaVA-1.6~\citep{llava, llava_next}.
When compared with other SoTA models, our \method{} models achieve competitive performance, despite being trained on much fewer and publicly available data only, and using a simple supervised fine-tuning training strategy.


Lastly, we present qualitative comparisons between our \method{} model and its S-Attn counterpart in Figure~\ref{fig:qualitative}.
We observe that the \method{} model provides answers that are more faithful to the input image and offers more visual details compared to the S-Attn model.
For example, in the first figure illustrating snowboarding and skiing, \method{} effectively distinguishes between the two activities, accurately identifying one person as skiing and the other as snowboarding.
While in the third figure at ``Diamond Head", \method{} provides more details about the scene such as ``encircled by a road that winds its way around the base'', and ``Beyond the crater, the city of Honolulu sprawls out''.


\subsection{Ablations and Analysis}\label{sec:ablation}

\noindent\textbf{V2V Diagonal-Attn.} We first conduct ablation studies on the V2V Diagonal-Attn, as detailed in Table~\ref{tab:ablation-arch}.
To demonstrate the computational advantages, we measure the maximum number of visual tokens ($|V|$) that an LVLM can process during training before encountering a GPU out-of-memory error.
We also record the training speed in seconds per iteration (sec/it) with the same $|V|$.
As shown in Table~\ref{tab:ablation-arch}, by diagonalizing the V2V Self-Attn, our model can process up to 8 times ($9k\rightarrow74k$) more visual tokens or train up to 5 times ($11.25s \rightarrow2.24s$) faster.
While additional optimization techniques such as FlashAttention~\citep{dao2022flashattention}, DeepSpeed~\citep{rasley2020deepspeed}, or Megatron~\citep{shoeybi2019megatron} can further improve memory and speed, they are orthogonal to our V2V Diagonal-Attn and still fundamentally have a computational complexity of $\mathcal{O}(|V|^2)$ for the V2V attention.
In terms of performance, V2V Diagonal-Attn performs comparably to conventional LVLMs across various benchmarks, supporting our hypothesis that visual tokens have already encoded contextual information, obviating the need for re-learning via the LLM's Self-Attn.


\noindent\textbf{Debiased Positional Encoding.} Next, we perform an ablation study on debiased positional encodings, also reported in Table~\ref{tab:ablation-arch}.
By debiasing the T2V Cross-Attn, our \method{} model achieves consistent performance improvements over models with biased positional encodings across multiple image benchmarks.
This modification cannot be easily implemented in conventional LVLMs but is rather straightforward with our proposed attention decomposition, and it brings no additional computational costs.


\begin{figure*}
    \centering
    \includegraphics[width=1.\linewidth]{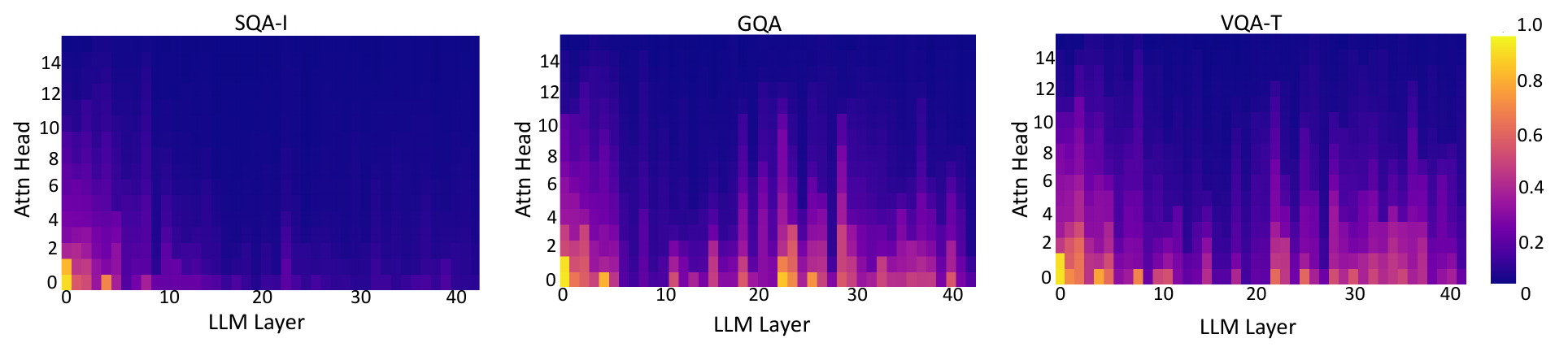}
    \vspace{-0.8cm}
    \caption{Visualization of $\alpha_{V}$ value on SQA-I, GQA, VQA-T benchmarks.}
    \label{fig:alpha_weight}
    \vspace{-0.2cm}
\end{figure*}
\begin{table*}[!htbp]
\centering
\begin{tabular}{@{\extracolsep{4pt}}c|cccccccccc@{}}
\toprule
Model & Existence & Count & Position & Color & Posters & Celebrity & Scene & Landmark & Artwork & OCR \\
\midrule
S-Attn & 190.0 & 165.0 & 121.7 & 180.0 & 134.4 & 161.2 & 166.3 & 157.8 & 128.0 & 102.5 \\
D-Attn & 195.0 & 170.0 & 143.3 & 195.0 & 161.6 & 172.6 & 163.0 & 166.8 & 137.0 & 132.5 \\
\bottomrule
\end{tabular}
\caption{
Detailed results of D-Attn and S-Attn on MME~\citep{fu2024mmecomprehensiveevaluationbenchmark} benchmark.
}\label{tab:mme}
\end{table*}
\begin{table*}[!htbp]
\centering
\renewcommand{\arraystretch}{1.2}
\resizebox{1.\linewidth}{!}{
\begin{tabular}{@{\extracolsep{4pt}}c|ccccccccc@{}}
\toprule
Model & \makecell{Scene\\Understanding} & \makecell{Instance\\Identity} & \makecell{Instance\\Location} & \makecell{Instance\\Attributes} & \makecell{Instances\\Counting} & \makecell{Spatial\\Relation} & \makecell{Instance\\Interaction} & \makecell{Visual\\Reasoning} & \makecell{Text\\Understanding} \\
\midrule
S-Attn & 76.9 & 74.5 & 74.7 & 67.3 & 64.2 & 57.8 & 73.2 & 76.1 & 44.7 \\
D-Attn & 78.1 & 78.2 & 77.5 & 68.6 & 67.5 & 61.0 & 73.2 & 80.9 & 65.8 \\
\bottomrule
\end{tabular}
}
\vspace{-0.2cm}
\caption{
Detailed results of D-Attn and S-Attn on SEED~\citep{li2023seed} benchmark.
}\label{tab:seed}
\vspace{-0.2cm}
\end{table*}
\begin{table*}[!htbp]
\centering
\renewcommand{\arraystretch}{1.2}
\resizebox{1.0\linewidth}{!}{
\begin{tabular}{@{\extracolsep{4pt}}c|ccccccccccc@{}}
\toprule
Model & \makecell{Action\\Recognition}  & \makecell{Attribute\\Recognition} & \makecell{Celebrity\\Recognition} & \makecell{Function\\Reasoning} & \makecell{Nature\\Relation} & \makecell{Object\\Localization} & Ocr & \makecell{Social\\Relation} & \makecell{Spatial\\Relationship} & \makecell{Struct. img-txt\\Understanding} \\
\midrule
S-Attn & 88.8 & 83.7 & 78.7 & 74.6 & 70.8 & 50.6 & 66.6 & 83.7 & 28.8 & 33.3 \\
D-Attn & 90.7 & 89.1 & 87.8 & 82.2 & 83.3 & 60.4 & 69.2 & 95.3 & 37.7 & 51.2 \\
\bottomrule
\end{tabular}
}
\vspace{-0.2cm}
\caption{
Detailed results of D-Attn and S-Attn on MMB~\citep{liu2023mmbench} benchmark.
}\label{tab:mmb}
\end{table*}

\noindent\textbf{Merging Strategies.} Furthermore, we experiment with different merging strategies in Table~\ref{tab:ablation-merge}, including (1) \textbf{Cascade}, where the T2V Cross-Attn module is decoupled and cascaded with T2T Self-Attn; (2) \textbf{Tanh}, where T2V Cross-Attn is weighted by a learnable tanh gate and then summed with T2T Self-Attn; (3) \textbf{Sigmoid}, where T2V Cross-Attn and T2T Self-Attn are weighted summed with learnable gates $\sigma$ and $1-\sigma$, respectively; and (4) $\bm{\alpha}$-\textbf{weighting} strategy proposed in this paper.
As shown in Table~\ref{tab:ablation-merge}, our $\alpha$-weighting strategy achieves superior performance compared to other strategies without introducing additional parameters like the cascade strategy.
Since $\alpha$-weighting introduces minimal architectural and operational changes to an LLM's self-attention module, it maximally retains the LLM's pre-trained capabilities, likely leading to superior fine-tuning performance on downstream tasks.

In Figure~\ref{fig:alpha_weight}, we further visualize the actual $\alpha_{V}$ values on these benchmarks (SQA-I, GQA, VQA-T) across attention heads and LLM layers. For each layer, we sort $\alpha_{V}$ across heads for better visualization. We can see that GQA and VQA-T both have high $\alpha_{V}$ across heads and layers, while SQA-I has much lower $\alpha_{V}$. This observation is in consensus with MM-Star~\citep{chen2024we}, \ie, many questions in SQA do not require visual information to answer.

\noindent\textbf{Benchmark Analysis.} Lastly, to gain deeper insights into the tasks that benefit most from our proposed \method{} model, we present the detailed scores for MME~\citep{fu2024mmecomprehensiveevaluationbenchmark}, SEED~\citep{li2023seed}, and MMB~\citep{liu2023mmbench} in Table~\ref{tab:mme},~\ref{tab:seed},~\ref{tab:mmb} respectively.
Our analysis reveals that our \method{} model excels particularly in tasks requiring spatial and relational reasoning.
Notable examples include (1) "position" in MME, (2) "Spatial Relation" in SEED, and (3) "object localization" and "spatial relationship" in MMB.
In addition, our \method{} model demonstrates strong performance on tasks involving OCR and document understanding.
Specific examples include (1) "OCR" in MME, (2) "Text Understanding" in SEED, and (3) "ocr" and "structuralized image-text understanding" in MMB.

\section{Conclusion}
\label{sec:conclusion}

In this paper, we introduce Decomposed Attention (D-Attn), a novel and flexible framework designed to process visual and textual tokens differently within LVLMs. Original self-attention is decomposed into three parts, \ie visual-to-visual, visual-to-textual, and textual-to-textual attentions. The visual and textual information are then merged back via the $\alpha$-weighting strategy to preserve the capabilities of pre-trained LLMs with minimal modifications. Furthermore, \method{} improves model performance by debiasing positional encoding in T2V Cross-Attn. In addition, \method{} reduces the computational complexity from $\mathcal{O}(|V|^2)$ to $\mathcal{O}(|V|)$ by diagonalizing V2V Self-Attn. Extensive experiments and rigorous analysis demonstrate that \method{} consistently outperforms its S-Attn counterpart, offering both performance gains and substantial computational savings.
Our contributions highlight the importance of handling visual and textual input separately with more flexibility, paving the way for more efficient and effective LVLMs.

{
    \small
    \bibliographystyle{ieeenat_fullname}
    \bibliography{main}

\begin{thebibliography}{45}
\providecommand{\natexlab}[1]{#1}
\providecommand{\url}[1]{\texttt{#1}}
\expandafter\ifx\csname urlstyle\endcsname\relax
  \providecommand{\doi}[1]{doi: #1}\else
  \providecommand{\doi}{doi: \begingroup \urlstyle{rm}\Url}\fi

\bibitem[Abdin et~al.(2024)Abdin, Jacobs, Awan, Aneja, Awadallah, Awadalla, Bach, Bahree, Bakhtiari, Behl, et~al.]{abdin2024phi}
Marah Abdin, Sam~Ade Jacobs, Ammar~Ahmad Awan, Jyoti Aneja, Ahmed Awadallah, Hany Awadalla, Nguyen Bach, Amit Bahree, Arash Bakhtiari, Harkirat Behl, et~al.
\newblock Phi-3 technical report: A highly capable language model locally on your phone.
\newblock \emph{arXiv preprint arXiv:2404.14219}, 2024.

\bibitem[Alayrac et~al.(2022)Alayrac, Donahue, Luc, Miech, Barr, Hasson, Lenc, Mensch, Millican, Reynolds, et~al.]{flamingo}
Jean-Baptiste Alayrac, Jeff Donahue, Pauline Luc, Antoine Miech, Iain Barr, Yana Hasson, Karel Lenc, Arthur Mensch, Katherine Millican, Malcolm Reynolds, et~al.
\newblock Flamingo: a visual language model for few-shot learning.
\newblock \emph{Advances in neural information processing systems}, 35:\penalty0 23716--23736, 2022.

\bibitem[Awadalla et~al.(2023)Awadalla, Gao, Gardner, Hessel, Hanafy, Zhu, Marathe, Bitton, Gadre, Sagawa, et~al.]{awadalla2023openflamingo}
Anas Awadalla, Irena Gao, Josh Gardner, Jack Hessel, Yusuf Hanafy, Wanrong Zhu, Kalyani Marathe, Yonatan Bitton, Samir Gadre, Shiori Sagawa, et~al.
\newblock Openflamingo: An open-source framework for training large autoregressive vision-language models.
\newblock \emph{arXiv preprint arXiv:2308.01390}, 2023.

\bibitem[Bai et~al.(2022)Bai, Jones, Ndousse, Askell, Chen, DasSarma, Drain, Fort, Ganguli, Henighan, et~al.]{bai2022training}
Yuntao Bai, Andy Jones, Kamal Ndousse, Amanda Askell, Anna Chen, Nova DasSarma, Dawn Drain, Stanislav Fort, Deep Ganguli, Tom Henighan, et~al.
\newblock Training a helpful and harmless assistant with reinforcement learning from human feedback.
\newblock \emph{arXiv preprint arXiv:2204.05862}, 2022.

\bibitem[Chen et~al.(2023)Chen, Li, Dong, Zhang, He, Wang, Zhao, and Lin]{chen2023sharegpt4v}
Lin Chen, Jisong Li, Xiaoyi Dong, Pan Zhang, Conghui He, Jiaqi Wang, Feng Zhao, and Dahua Lin.
\newblock Sharegpt4v: Improving large multi-modal models with better captions.
\newblock \emph{arXiv preprint arXiv:2311.12793}, 2023.

\bibitem[Chen et~al.(2024)Chen, Li, Dong, Zhang, Zang, Chen, Duan, Wang, Qiao, Lin, et~al.]{chen2024we}
Lin Chen, Jinsong Li, Xiaoyi Dong, Pan Zhang, Yuhang Zang, Zehui Chen, Haodong Duan, Jiaqi Wang, Yu Qiao, Dahua Lin, et~al.
\newblock Are we on the right way for evaluating large vision-language models?
\newblock \emph{arXiv preprint arXiv:2403.20330}, 2024.

\bibitem[Dai et~al.(2023)Dai, Li, Li, Tiong, Zhao, Wang, Li, Fung, and Hoi]{dai2023instructblip}
Wenliang Dai, Junnan Li, Dongxu Li, Anthony Tiong, Junqi Zhao, Weisheng Wang, Boyang Li, Pascale Fung, and Steven Hoi.
\newblock Instruct{BLIP}: Towards general-purpose vision-language models with instruction tuning.
\newblock In \emph{Thirty-seventh Conference on Neural Information Processing Systems}, 2023.

\bibitem[Dao et~al.(2022)Dao, Fu, Ermon, Rudra, and R{\'e}]{dao2022flashattention}
Tri Dao, Dan Fu, Stefano Ermon, Atri Rudra, and Christopher R{\'e}.
\newblock Flashattention: Fast and memory-efficient exact attention with io-awareness.
\newblock \emph{Advances in Neural Information Processing Systems}, 35:\penalty0 16344--16359, 2022.

\bibitem[Deitke et~al.(2024)Deitke, Clark, Lee, Tripathi, Yang, Park, Salehi, Muennighoff, Lo, Soldaini, et~al.]{deitke2024molmo}
Matt Deitke, Christopher Clark, Sangho Lee, Rohun Tripathi, Yue Yang, Jae~Sung Park, Mohammadreza Salehi, Niklas Muennighoff, Kyle Lo, Luca Soldaini, et~al.
\newblock Molmo and pixmo: Open weights and open data for state-of-the-art multimodal models.
\newblock \emph{arXiv preprint arXiv:2409.17146}, 2024.

\bibitem[Dubey et~al.(2024)Dubey, Jauhri, Pandey, Kadian, Al-Dahle, Letman, Mathur, Schelten, Yang, Fan, et~al.]{dubey2024llama}
Abhimanyu Dubey, Abhinav Jauhri, Abhinav Pandey, Abhishek Kadian, Ahmad Al-Dahle, Aiesha Letman, Akhil Mathur, Alan Schelten, Amy Yang, Angela Fan, et~al.
\newblock The llama 3 herd of models.
\newblock \emph{arXiv preprint arXiv:2407.21783}, 2024.

\bibitem[Fu et~al.(2024)Fu, Chen, Shen, Qin, Zhang, Lin, Yang, Zheng, Li, Sun, Wu, and Ji]{fu2024mmecomprehensiveevaluationbenchmark}
Chaoyou Fu, Peixian Chen, Yunhang Shen, Yulei Qin, Mengdan Zhang, Xu Lin, Jinrui Yang, Xiawu Zheng, Ke Li, Xing Sun, Yunsheng Wu, and Rongrong Ji.
\newblock Mme: A comprehensive evaluation benchmark for multimodal large language models, 2024.

\bibitem[Goyal et~al.(2017)Goyal, Khot, Summers-Stay, Batra, and Parikh]{goyal2017making}
Yash Goyal, Tejas Khot, Douglas Summers-Stay, Dhruv Batra, and Devi Parikh.
\newblock Making the v in vqa matter: Elevating the role of image understanding in visual question answering.
\newblock In \emph{Proceedings of the IEEE conference on computer vision and pattern recognition}, pages 6904--6913, 2017.

\bibitem[Hudson and Manning(2019)]{hudson2019gqa}
Drew~A Hudson and Christopher~D Manning.
\newblock Gqa: A new dataset for real-world visual reasoning and compositional question answering.
\newblock In \emph{Proceedings of the IEEE/CVF conference on computer vision and pattern recognition}, pages 6700--6709, 2019.

\bibitem[Jiang et~al.(2023)Jiang, Sablayrolles, Mensch, Bamford, Chaplot, Casas, Bressand, Lengyel, Lample, Saulnier, et~al.]{jiang2023mistral}
Albert~Q Jiang, Alexandre Sablayrolles, Arthur Mensch, Chris Bamford, Devendra~Singh Chaplot, Diego de~las Casas, Florian Bressand, Gianna Lengyel, Guillaume Lample, Lucile Saulnier, et~al.
\newblock Mistral 7b.
\newblock \emph{arXiv preprint arXiv:2310.06825}, 2023.

\bibitem[Lauren{\c{c}}on et~al.(2024{\natexlab{a}})Lauren{\c{c}}on, Saulnier, Tronchon, Bekman, Singh, Lozhkov, Wang, Karamcheti, Rush, Kiela, et~al.]{laurenccon2024obelics}
Hugo Lauren{\c{c}}on, Lucile Saulnier, L{\'e}o Tronchon, Stas Bekman, Amanpreet Singh, Anton Lozhkov, Thomas Wang, Siddharth Karamcheti, Alexander Rush, Douwe Kiela, et~al.
\newblock Obelics: An open web-scale filtered dataset of interleaved image-text documents.
\newblock \emph{Advances in Neural Information Processing Systems}, 36, 2024{\natexlab{a}}.

\bibitem[Lauren{\c{c}}on et~al.(2024{\natexlab{b}})Lauren{\c{c}}on, Tronchon, Cord, and Sanh]{laurenccon2024matters}
Hugo Lauren{\c{c}}on, L{\'e}o Tronchon, Matthieu Cord, and Victor Sanh.
\newblock What matters when building vision-language models?
\newblock \emph{arXiv preprint arXiv:2405.02246}, 2024{\natexlab{b}}.

\bibitem[Li et~al.(2023)Li, Wang, Wang, Ge, Ge, and Shan]{li2023seed}
Bohao Li, Rui Wang, Guangzhi Wang, Yuying Ge, Yixiao Ge, and Ying Shan.
\newblock Seed-bench: Benchmarking multimodal llms with generative comprehension.
\newblock \emph{arXiv preprint arXiv:2307.16125}, 2023.

\bibitem[Li et~al.(2024{\natexlab{a}})Li, Zhang, Guo, Zhang, Li, Zhang, Zhang, Li, Liu, and Li]{llava_onevision}
Bo Li, Yuanhan Zhang, Dong Guo, Renrui Zhang, Feng Li, Hao Zhang, Kaichen Zhang, Yanwei Li, Ziwei Liu, and Chunyuan Li.
\newblock Llava-onevision: Easy visual task transfer.
\newblock \emph{arXiv preprint arXiv:2408.03326}, 2024{\natexlab{a}}.

\bibitem[Li et~al.(2024{\natexlab{b}})Li, Wang, Zhu, Kuo, Xu, Chen, Jain, Shi, and Wen]{li2024cumo}
Jiachen Li, Xinyao Wang, Sijie Zhu, Chia-Wen Kuo, Lu Xu, Fan Chen, Jitesh Jain, Humphrey Shi, and Longyin Wen.
\newblock Cumo: Scaling multimodal llm with co-upcycled mixture-of-experts.
\newblock \emph{arXiv preprint arXiv:2405.05949}, 2024{\natexlab{b}}.

\bibitem[Li et~al.(2024{\natexlab{c}})Li, Zhang, Wang, Zhong, Chen, Chu, Liu, and Jia]{li2024mini}
Yanwei Li, Yuechen Zhang, Chengyao Wang, Zhisheng Zhong, Yixin Chen, Ruihang Chu, Shaoteng Liu, and Jiaya Jia.
\newblock Mini-gemini: Mining the potential of multi-modality vision language models.
\newblock \emph{arXiv preprint arXiv:2403.18814}, 2024{\natexlab{c}}.

\bibitem[Lin et~al.(2024)Lin, Yin, Ping, Molchanov, Shoeybi, and Han]{lin2024vila}
Ji Lin, Hongxu Yin, Wei Ping, Pavlo Molchanov, Mohammad Shoeybi, and Song Han.
\newblock Vila: On pre-training for visual language models.
\newblock In \emph{Proceedings of the IEEE/CVF Conference on Computer Vision and Pattern Recognition}, pages 26689--26699, 2024.

\bibitem[Liu et~al.(2024{\natexlab{a}})Liu, Li, Li, Li, Zhang, Shen, and Lee]{llava_next}
Haotian Liu, Chunyuan Li, Yuheng Li, Bo Li, Yuanhan Zhang, Sheng Shen, and Yong~Jae Lee.
\newblock Llava-next: Improved reasoning, ocr, and world knowledge, 2024{\natexlab{a}}.

\bibitem[Liu et~al.(2024{\natexlab{b}})Liu, Li, Wu, and Lee]{llava}
Haotian Liu, Chunyuan Li, Qingyang Wu, and Yong~Jae Lee.
\newblock Visual instruction tuning.
\newblock \emph{Advances in neural information processing systems}, 36, 2024{\natexlab{b}}.

\bibitem[Liu et~al.(2023)Liu, Duan, Zhang, Li, Zhang, Zhao, Yuan, Wang, He, Liu, et~al.]{liu2023mmbench}
Yuan Liu, Haodong Duan, Yuanhan Zhang, Bo Li, Songyang Zhang, Wangbo Zhao, Yike Yuan, Jiaqi Wang, Conghui He, Ziwei Liu, et~al.
\newblock Mmbench: Is your multi-modal model an all-around player?
\newblock \emph{arXiv preprint arXiv:2307.06281}, 2023.

\bibitem[Lu et~al.(2022)Lu, Mishra, Xia, Qiu, Chang, Zhu, Tafjord, Clark, and Kalyan]{lu2022learn}
Pan Lu, Swaroop Mishra, Tanglin Xia, Liang Qiu, Kai-Wei Chang, Song-Chun Zhu, Oyvind Tafjord, Peter Clark, and Ashwin Kalyan.
\newblock Learn to explain: Multimodal reasoning via thought chains for science question answering.
\newblock \emph{Advances in Neural Information Processing Systems}, 35:\penalty0 2507--2521, 2022.

\bibitem[Mao et~al.(2016)Mao, Huang, Toshev, Camburu, Yuille, and Murphy]{mao2016generation}
Junhua Mao, Jonathan Huang, Alexander Toshev, Oana Camburu, Alan~L Yuille, and Kevin Murphy.
\newblock Generation and comprehension of unambiguous object descriptions.
\newblock In \emph{Proceedings of the IEEE conference on computer vision and pattern recognition}, pages 11--20, 2016.

\bibitem[Ordonez et~al.(2011)Ordonez, Kulkarni, and Berg]{ordonez2011im2text}
Vicente Ordonez, Girish Kulkarni, and Tamara Berg.
\newblock Im2text: Describing images using 1 million captioned photographs.
\newblock \emph{Advances in neural information processing systems}, 24, 2011.

\bibitem[Radford et~al.(2021)Radford, Kim, Hallacy, Ramesh, Goh, Agarwal, Sastry, Askell, Mishkin, Clark, et~al.]{clip}
Alec Radford, Jong~Wook Kim, Chris Hallacy, Aditya Ramesh, Gabriel Goh, Sandhini Agarwal, Girish Sastry, Amanda Askell, Pamela Mishkin, Jack Clark, et~al.
\newblock Learning transferable visual models from natural language supervision.
\newblock In \emph{International conference on machine learning}, pages 8748--8763. PMLR, 2021.

\bibitem[Rafailov et~al.(2024)Rafailov, Sharma, Mitchell, Manning, Ermon, and Finn]{rafailov2024direct}
Rafael Rafailov, Archit Sharma, Eric Mitchell, Christopher~D Manning, Stefano Ermon, and Chelsea Finn.
\newblock Direct preference optimization: Your language model is secretly a reward model.
\newblock \emph{Advances in Neural Information Processing Systems}, 36, 2024.

\bibitem[Rasley et~al.(2020)Rasley, Rajbhandari, Ruwase, and He]{rasley2020deepspeed}
Jeff Rasley, Samyam Rajbhandari, Olatunji Ruwase, and Yuxiong He.
\newblock Deepspeed: System optimizations enable training deep learning models with over 100 billion parameters.
\newblock In \emph{Proceedings of the 26th ACM SIGKDD International Conference on Knowledge Discovery \& Data Mining}, pages 3505--3506, 2020.

\bibitem[Schuhmann et~al.(2022)Schuhmann, Beaumont, Vencu, Gordon, Wightman, Cherti, Coombes, Katta, Mullis, Wortsman, et~al.]{schuhmann2022laion}
Christoph Schuhmann, Romain Beaumont, Richard Vencu, Cade Gordon, Ross Wightman, Mehdi Cherti, Theo Coombes, Aarush Katta, Clayton Mullis, Mitchell Wortsman, et~al.
\newblock Laion-5b: An open large-scale dataset for training next generation image-text models.
\newblock \emph{Advances in Neural Information Processing Systems}, 35:\penalty0 25278--25294, 2022.

\bibitem[Sharma et~al.(2018)Sharma, Ding, Goodman, and Soricut]{sharma2018conceptual}
Piyush Sharma, Nan Ding, Sebastian Goodman, and Radu Soricut.
\newblock Conceptual captions: A cleaned, hypernymed, image alt-text dataset for automatic image captioning.
\newblock In \emph{Proceedings of ACL}, 2018.

\bibitem[Shoeybi et~al.(2019)Shoeybi, Patwary, Puri, LeGresley, Casper, and Catanzaro]{shoeybi2019megatron}
Mohammad Shoeybi, Mostofa Patwary, Raul Puri, Patrick LeGresley, Jared Casper, and Bryan Catanzaro.
\newblock Megatron-lm: Training multi-billion parameter language models using model parallelism.
\newblock \emph{arXiv preprint arXiv:1909.08053}, 2019.

\bibitem[Team(2024)]{team2024chameleon}
Chameleon Team.
\newblock Chameleon: Mixed-modal early-fusion foundation models.
\newblock \emph{arXiv preprint arXiv:2405.09818}, 2024.

\bibitem[Team et~al.(2024)Team, Riviere, Pathak, Sessa, Hardin, Bhupatiraju, Hussenot, Mesnard, Shahriari, Ram{\'e}, et~al.]{team2024gemma}
Gemma Team, Morgane Riviere, Shreya Pathak, Pier~Giuseppe Sessa, Cassidy Hardin, Surya Bhupatiraju, L{\'e}onard Hussenot, Thomas Mesnard, Bobak Shahriari, Alexandre Ram{\'e}, et~al.
\newblock Gemma 2: Improving open language models at a practical size.
\newblock \emph{arXiv preprint arXiv:2408.00118}, 2024.

\bibitem[Tong et~al.(2024{\natexlab{a}})Tong, Brown, Wu, Woo, Middepogu, Akula, Yang, Yang, Iyer, Pan, et~al.]{tong2024cambrian}
Shengbang Tong, Ellis Brown, Penghao Wu, Sanghyun Woo, Manoj Middepogu, Sai~Charitha Akula, Jihan Yang, Shusheng Yang, Adithya Iyer, Xichen Pan, et~al.
\newblock Cambrian-1: A fully open, vision-centric exploration of multimodal llms.
\newblock \emph{arXiv preprint arXiv:2406.16860}, 2024{\natexlab{a}}.

\bibitem[Tong et~al.(2024{\natexlab{b}})Tong, Liu, Zhai, Ma, LeCun, and Xie]{tong2024eyes}
Shengbang Tong, Zhuang Liu, Yuexiang Zhai, Yi Ma, Yann LeCun, and Saining Xie.
\newblock Eyes wide shut? exploring the visual shortcomings of multimodal llms.
\newblock In \emph{Proceedings of the IEEE/CVF Conference on Computer Vision and Pattern Recognition}, pages 9568--9578, 2024{\natexlab{b}}.

\bibitem[Touvron et~al.(2023)Touvron, Lavril, Izacard, Martinet, Lachaux, Lacroix, Rozi{\`e}re, Goyal, Hambro, Azhar, et~al.]{touvron2023llama}
Hugo Touvron, Thibaut Lavril, Gautier Izacard, Xavier Martinet, Marie-Anne Lachaux, Timoth{\'e}e Lacroix, Baptiste Rozi{\`e}re, Naman Goyal, Eric Hambro, Faisal Azhar, et~al.
\newblock Llama: Open and efficient foundation language models.
\newblock \emph{arXiv preprint arXiv:2302.13971}, 2023.

\bibitem[Vaswani(2017)]{vaswani2017attention}
A Vaswani.
\newblock Attention is all you need.
\newblock \emph{Advances in Neural Information Processing Systems}, 2017.

\bibitem[Wang et~al.(2024)Wang, Bai, Tan, Wang, Fan, Bai, Chen, Liu, Wang, Ge, et~al.]{wang2024qwen2}
Peng Wang, Shuai Bai, Sinan Tan, Shijie Wang, Zhihao Fan, Jinze Bai, Keqin Chen, Xuejing Liu, Jialin Wang, Wenbin Ge, et~al.
\newblock Qwen2-vl: Enhancing vision-language model's perception of the world at any resolution.
\newblock \emph{arXiv preprint arXiv:2409.12191}, 2024.

\bibitem[Xue et~al.(2024)Xue, Shu, Awadalla, Wang, Yan, Purushwalkam, Zhou, Prabhu, Dai, Ryoo, et~al.]{blip3}
Le Xue, Manli Shu, Anas Awadalla, Jun Wang, An Yan, Senthil Purushwalkam, Honglu Zhou, Viraj Prabhu, Yutong Dai, Michael~S Ryoo, et~al.
\newblock xgen-mm (blip-3): A family of open large multimodal models.
\newblock \emph{arXiv preprint arXiv:2408.08872}, 2024.

\bibitem[Zhai et~al.(2023)Zhai, Mustafa, Kolesnikov, and Beyer]{zhai2023sigmoid}
Xiaohua Zhai, Basil Mustafa, Alexander Kolesnikov, and Lucas Beyer.
\newblock Sigmoid loss for language image pre-training.
\newblock In \emph{Proceedings of the IEEE/CVF International Conference on Computer Vision}, pages 11975--11986, 2023.

\bibitem[Zhang and Sennrich(2019)]{zhang2019root}
Biao Zhang and Rico Sennrich.
\newblock Root mean square layer normalization.
\newblock \emph{Advances in Neural Information Processing Systems}, 32, 2019.

\bibitem[Zhang et~al.(2024)Zhang, Dong, Zang, Cao, Qian, Chen, Guo, Duan, Wang, Ouyang, et~al.]{zhang2024internlm}
Pan Zhang, Xiaoyi Dong, Yuhang Zang, Yuhang Cao, Rui Qian, Lin Chen, Qipeng Guo, Haodong Duan, Bin Wang, Linke Ouyang, et~al.
\newblock Internlm-xcomposer-2.5: A versatile large vision language model supporting long-contextual input and output.
\newblock \emph{arXiv preprint arXiv:2407.03320}, 2024.

\bibitem[Zheng et~al.(2023)Zheng, Chiang, Sheng, Zhuang, Wu, Zhuang, Lin, Li, Li, Xing, et~al.]{zheng2023judging}
Lianmin Zheng, Wei-Lin Chiang, Ying Sheng, Siyuan Zhuang, Zhanghao Wu, Yonghao Zhuang, Zi Lin, Zhuohan Li, Dacheng Li, Eric Xing, et~al.
\newblock Judging llm-as-a-judge with mt-bench and chatbot arena.
\newblock \emph{Advances in Neural Information Processing Systems}, 36:\penalty0 46595--46623, 2023.

\end{thebibliography}
}


\end{document}